\pdfoutput=1

\documentclass[11pt]{article}

\usepackage[preprint]{acl}
\usepackage{times}
\usepackage{latexsym}

\usepackage[T1]{fontenc}

\usepackage[utf8]{inputenc}

\usepackage{microtype}

\usepackage{graphicx}
\usepackage{inconsolata}
\usepackage{algorithm}
\usepackage{algpseudocode}
\usepackage{algorithmicx}
\usepackage{threeparttable}
\usepackage{multirow}
\usepackage{multicol}
\usepackage{makecell}
\usepackage{booktabs}
\usepackage{amsmath,bm}
\usepackage{amsfonts}
\usepackage{bbding}
\usepackage{pifont}

\newcommand*{\affaddr}[1]{#1} 
\newcommand*{\email}[1]{\texttt{#1}}

\title{Multi-Prompting Decoder Helps Better Language Understanding}

\author{%
{Zifeng Cheng\thanks{\ \ Both authors contributed equally to this research.}, Zhaoling Chen\footnotemark[1], Zhiwei Jiang\thanks{\ \ Corresponding author.}, Yafeng Yin,}\\ 
\textbf{Cong Wang, Shiping Ge, Qing Gu}\\
\affaddr{National Key Laboratory for Novel Software Technology, Nanjing University}\\
\email{\{chengzf,zhaolingchen\}@smail.nju.edu.cn}\qquad
\email{\{jzw,yafeng\}@nju.edu.cn}\\
\email{\{cw,shipingge\}@smail.nju.edu.cn} \qquad
\email{guq@nju.edu.cn}%
}

\begin{document}
\maketitle
\begin{abstract}
Recent large Pre-trained Language Models (PLMs) usually only provide users with the inference APIs, namely the emerging Model-as-a-Service (MaaS) setting.
To adapt MaaS PLMs to downstream tasks without accessing their parameters and gradients, some existing methods focus on the output-side adaptation of PLMs, viewing the PLM as an encoder and then optimizing a task-specific decoder for decoding the output hidden states and class scores of the PLM.
Despite the effectiveness of these methods, they only use a single prompt to query PLMs for decoding, leading to a heavy reliance on the quality of the adopted prompt.
In this paper, we propose a simple yet effective Multi-Prompting Decoder (MPD) framework for MaaS adaptation.
The core idea is to query PLMs with multiple different prompts for each sample, thereby obtaining multiple output hidden states and class scores from PLMs for subsequent decoding.
Such multi-prompting decoding paradigm can simultaneously mitigate reliance on the quality of a single prompt, alleviate the issue of data scarcity under the few-shot setting, and provide richer knowledge extracted from PLMs.
Specifically, we propose two decoding strategies: multi-prompting decoding with optimal transport for hidden states and calibrated decoding for class scores.
Extensive experiments demonstrate that our method is effective on multiple natural language understanding datasets under the few-shot setting.
\end{abstract}

\section{Introduction}
Pre-trained Language Models (PLMs) have recently achieved remarkable performance on various downstream language understanding tasks with the ``pre-training and fine-tuning'' paradigm~\cite{DBLP:journals/csur/LiuYFJHN23}.
However, as the scale of PLMs continues to grow, fine-tuning the full model for each downstream task has become computationally expensive and deployment-inefficient.
In light of this, Model-as-a-Service (MaaS)~\cite{DBLP:journals/corr/abs-2201-08531Diao,DBLP:conf/icml/SunSQHQ22} has emerged as a more practical model deployment scheme, allowing pre-trained models to serve downstream tasks by providing inference APIs.
Within the MaaS scheme, users can access the outputs of PLMs (e.g., \textit{hidden states}, \textit{class scores}, or \textit{textual output}) through APIs\footnote{Currently, some famous MaaS services include GPT-3.5 Turbo, which offers APIs for accessing class scores and text output, as well as text-embedding-3-small and text-embedding-3-large, which provide APIs for accessing embedding representations (i.e., \textit{hidden states}) of input text.}, but cannot access the parameters and gradients of PLMs, presenting a challenge in effectively adapting PLMs to downstream tasks.

To adapt MaaS PLMs to downstream tasks, existing methods have primarily explored both the input-side and the output-side adaptation.
On the input-side adaptation, researchers focus on utilizing gradient-free methods to heuristically search a good continuous~\cite{DBLP:conf/icml/SunSQHQ22,DBLP:conf/emnlp/SunHQZHQ22} or discrete prompt~\cite{DBLP:journals/corr/abs-2201-08531Diao,DBLP:conf/emnlp/DengWHWGSSXH22} for the target task, such as using evolutionary algorithm and reinforcement learning.
However, input-side methods need to query PLM thousands of times for optimization, resulting in significant time overhead and the optimization process is very difficult due to the huge search space and the lack of gradients.
On the output-side adaptation, researchers \cite{hou2023promptboosting,DBLP:conf/acl/CuiLDHLS23} treat the PLM as an encoder and locally train a task-specific decoder to post-process PLM's outputs.
Compared to the input-side adaptation, the output-side adaptation methods only query PLM a few times and can benefit from gradient for optimization, often resulting in better performance.




\begin{figure}[t]
\centering
\includegraphics[width=1\columnwidth]{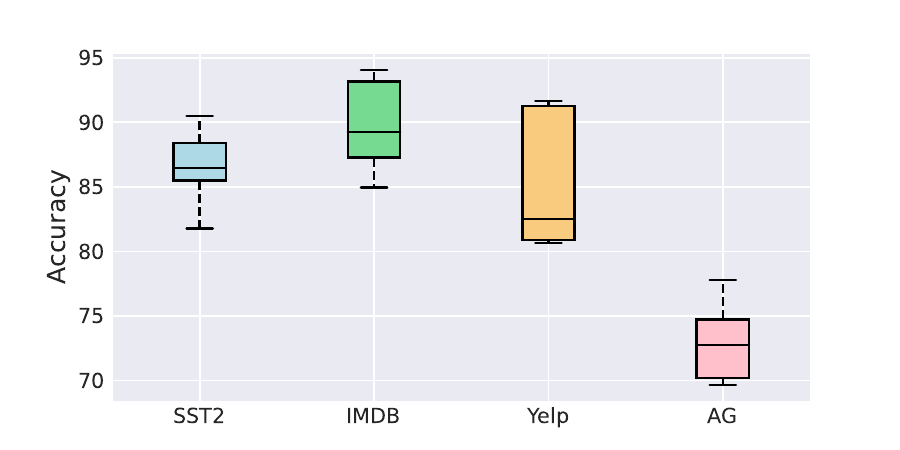}
\caption{Performance and variation using five different prompts on four datasets.} \label{fig:motivation}
\end{figure}

Despite the effectiveness of these output-side adaptation methods, they usually only adopt a single prompt to query PLMs for a certain downstream task, thus often struggling with the performance variations brought by using different prompts \cite{DBLP:conf/acl/LuBM0S22,DBLP:conf/acl/ChenDWH0LX23}.
As shown in Figure~\ref{fig:motivation}, on all four datasets, the maximum performance fluctuations in accuracy can exceed 8\% when using different prompts.
Therefore, the quality of prompt is crucial, and relying on only one prompt easily exposes them to performance risks of encountering a low-quality query.

To this end, we propose to query PLMs with multiple prompts for a certain downstream task. 
This multi-prompting paradigm offers several advantages. 
Firstly, using multiple prompts helps mitigate reliance on the quality of a single prompt, leading to better stability.
Secondly, in the few-shot setting, using multiple prompts enables a single sample to obtain multiple representations, thereby addressing the issue of data scarcity. 
Lastly, multiple prompts can guide the extraction of knowledge from PLMs from different perspectives, providing richer knowledge for downstream tasks.

Based on the above analysis, in this paper, we propose a novel output-side PLMs adaptation framework for few-shot classification, named Multi-Prompting Decoder (MPD).
This MPD framework can decode the outputs of PLMs, including hidden states and class scores, to yield classification results.
In contrast to previous methods, MPD adopts a novel multi-prompting decoding paradigm, which queries PLMs with multiple prompts for a sample, thereby obtaining more representation information for decoding.
Specifically, we propose two decoding strategies: multi-prompting decoding with optimal transport for hidden states and calibrated multi-prompting decoding for class scores.
Extensive experiments demonstrate that our method is effective under the few-shot setting on multiple natural language understanding tasks, including sentiment analysis, topic classification, and natural language inference.

\section{Related Work}

\textbf{Prompt Tuning} Prompt tuning aims to reduce the gap between PLM pre-training and fine-tuning via predicting \texttt{[MASK]} token in the prompt for data-efficiency and achieves great performance in zero-shot and few-shot learning \cite{DBLP:journals/csur/LiuYFJHN23}.
Since prompt has a large impact on the performance of prompt tuning, many studies have focused on finding a good prompt, including discrete prompt~\cite{DBLP:journals/tacl/JiangXAN20,DBLP:conf/emnlp/ShinRLWS20,DBLP:conf/acl/GaoFC20}, continuous prompt~\cite{DBLP:journals/corr/abs-2103-10385liu,DBLP:conf/naacl/QinE21,DBLP:conf/acl/LiL20,DBLP:conf/emnlp/LesterAC21,DBLP:conf/iclr/ZhangLCDBTHC22}, prompt selection~\cite{DBLP:conf/acl/SorensenRRSRDKF22,DBLP:conf/acl/ChenDWH0LX23}, and so on.

\textbf{MaaS Adaptation} Some PLMs, such as GPT-3 \cite{DBLP:conf/nips/BrownMRSKDNSSAA20}, are released as a service in the cloud.
Some works try to adapt these models to downstream tasks without accessing the model parameters and the gradients~\cite{DBLP:conf/icml/SunSQHQ22,DBLP:journals/corr/abs-2201-08531Diao}.

Some work about MaaS uses reinforcement learning (RL)~\cite{DBLP:journals/corr/abs-2201-08531Diao,DBLP:conf/emnlp/DengWHWGSSXH22} or derivative-free optimization~\cite{DBLP:conf/icml/SunSQHQ22,DBLP:conf/emnlp/SunHQZHQ22} to find the optimal prompt.
For the first type of method,~\citet{DBLP:journals/corr/abs-2201-08531Diao} first use a variance-reduced policy gradient algorithm to estimate the gradient of the prompt token distribution.
\citet{DBLP:conf/emnlp/DengWHWGSSXH22} use a policy network and reward function to generate and optimize discrete prompts.
For the second type of method, BBT \cite{DBLP:conf/icml/SunSQHQ22} adopts a covariance matrix adaptation evolution algorithm to optimize continuous prompt.
BBTv2 \cite{DBLP:conf/emnlp/SunHQZHQ22} further prepends continuous prompts to every layer of PLM and proposes a divide-and-conquer gradient-free algorithm to optimize them.
GDFO \cite{DBLP:conf/acl/Han23} uses a prompt generator to generate an extra vector as input based on the BBT framework.
TEMPERA~\cite{DBLP:conf/iclr/Zhang0ZSG23} constructs query-dependent prompts through test-time prompt editing and formulates this as an RL problem.
DP$_2$O~\cite{li2023dialogue} first designs a prompt generation strategy through multi-round dialogue alignment on GPT-4 and prompt evaluation metric SUE.
Then, DP$_2$O constructs RL framework based on policy gradients to match the prompts to inputs optimally.

However, searching in a large space of prompts is inefficient and difficult to optimize.
Inspired by this, PromptBoosting \cite{hou2023promptboosting} constructs a series of weak learners, learns the corresponding verbalizer, and uses AdaBoost algorithm to ensemble them.
DecT \cite{DBLP:conf/acl/CuiLDHLS23} is an output-side adaptation framework and optimizes class prototypes as hyperspheres with a radius parameter.

It's worth noting that parameter-efficient tuning \cite{DBLP:conf/icml/HoulsbyGJMLGAG19,DBLP:conf/iclr/HuSWALWWC22,DBLP:conf/acl/ZakenGR22,DBLP:conf/iclr/HeZMBN22} requires gradients of PLMs to update the small portion of parameters, which are unavailable in the MaaS setting.

\begin{figure*}[t]
\centering
\includegraphics[width=2\columnwidth]{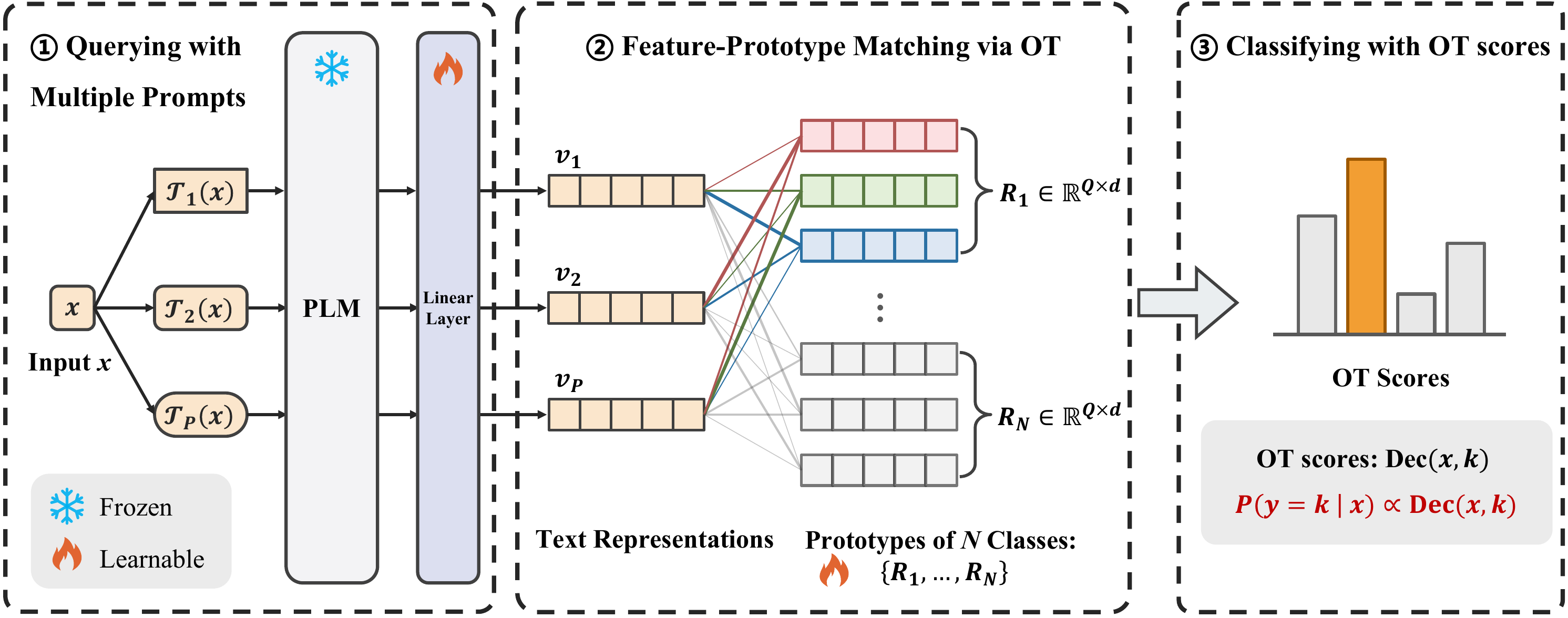}
\caption{Illustration of the multi-prompting decoding with optimal transport.} \label{fig:framework}
\end{figure*}

\textbf{Optimal Transport}
OT theory~\cite{monge1781memoire} was originally used to study the movement of many items from one place to another with minimal cost and provides a distance measure method between two distributions in a closed way.
Due to the excellent properties of the distance measure, OT is used in many areas, such as topic model~\cite{DBLP:conf/iclr/ZhaoPHLB21}, vision-language models~\cite{DBLP:conf/iclr/0002YSLR023}, and image classification~\cite{liu2021multi}.
                                                     
\section{Preliminaries} \label{sec:pre}

MaaS adaptation aims to correctly predict the label $y\in\{1,\dots, N\}$ for test sample $x$ based on a given few-shot training set $\mathcal{D}_{\text{train}} = \{ (x_i, y_i)\}_{i=1}^{NK}$ and PLM $\mathcal{M}$, where $N$ is the number of classes and each class has $K$ training samples.
In the MaaS setting, PLM $\mathcal{M}$ is a black-box inference API without accessing the model parameters and the gradients.
Therefore, we can only query the model with input $x$ and get corresponding outputs, including hidden states and class scores.

To better utilize the PLM, we use prompt learning to wrap input samples into prompts using templates.
Specifically, we first use a template $\mathcal{T}$ with a \texttt{[MASK]} token to enclose each input.
Then, we can query $\mathcal{M}$ with prompt $\mathcal{T}(x)$ to get the hidden states and class scores.
Hidden states can be obtained by the representation of \texttt{[MASK]} token of the final layer and class scores $\mathbf{s} \in \mathbb{R}^{N}$ can be obtained by using the verbalizer to map predicted tokens at the \texttt{[MASK]} token to classes.
Take sentiment analysis as an example, we can use 
$$\mathcal{T}(x) = x\ \text{In summary, it was \texttt{[MASK]}.}$$ as prompt with verbalizer that maps ``bad'' and ``great'' as label words for negative and positive sentiment.

\section{Methodology}

We introduce a novel output-side MaaS adaptation framework, Multi-Prompting Decoder (MPD), for few-shot classification.
This framework can decode the outputs of PLMs, including hidden states and class scores, to yield classification results.
In contrast to previous decoding frameworks that query PLMs only once for a sample, we propose multi-prompting decoding paradigm, which queries PLMs with multiple prompts for a sample, thereby obtaining more representation information for decoding.
Specifically, for decoding the PLMs' output hidden states, we design a multi-prompting decoding strategy based on optimal transport.
For decoding the PLMs' output class scores, we additionally design a calibrated multi-prompting decoding strategy.
These two decoding strategies are jointly used in the final decoding process.

\subsection{Multi-Prompting Decoding with Optimal Transport}
In our multi-prompting paradigm, each sample has multiple types of hidden states corresponding to multiple prompts. Building a unified classifier for these hidden states neglects the characteristics of each prompt, leading to a crude solution.
Building multiple classifiers specific to each type of hidden state is also a sub-optimal solution, which will lead to an ensemble of multiple classifiers, and these classifiers cannot benefit from each other during training.

To fully exploit the advantages of the multi-prompting decoding paradigm, we design a specific multi-prompting decoder for hidden states. 
The core idea is to establish multiple prototypes for each class to capture the class characteristics specific to different prompts. 
Subsequently, through optimal transport, we identify the best matching between the sample representations and the prototypes, thereby achieving classification. 
Considering the limited number of training samples available in the few-shot setting, the decoder is lightweight, comprising only a linear layer and several learnable class prototypes.

\subsubsection{Querying with Multiple Prompts}
Given a text $x_i$, we first use a series of templates $\mathcal{T}_1, \cdots, \mathcal{T}_P$ with a \texttt{[MASK]} token to enclose the text and then query $\mathcal{M}$ with template-wrapped inputs $\mathcal{T}_1(x_i)$, $\cdots$, $\mathcal{T}_P(x_i)$ to get a series of initial text representations $\mathbf{h}_{i,1}, \cdots, \mathbf{h}_{i,P}$.
Specifically, we take hidden states of the final layer at the \texttt{[MASK]} position as the initial text representations extracted by PLM $\mathcal{M}$.
Then, we use a linear layer parameterized by $\mathbf{W}$ to project the initial representations to get text representations for classification:
\begin{equation}
    \mathbf{v}  = \mathbf{Wh}
\end{equation}
It is worth noting that the text representation matrix of text $x_i$ is $\mathbf{V}_i \in \mathbb{R}^{P \times d}$.

\subsubsection{Features-Prototypes Matching via OT}

Since our framework contains multiple text representations and class prototypes, OT is a proper solution to better map multiple text representations and multiple prototypes of each class.
OT aims to find the optimal transport plan $\bm{T} \in \mathbb{R}^{P \times Q}$ that represents the fine-grained matching flow between text representations and class prototypes, where $P$ denotes the number of prompts and $Q$ denotes the number of prototypes.
The optimal transport plan assigns a higher weight to the pair of text representation and prototype that are close to each other.
We use Sinkhorn's algorithm \cite{cuturi2013sinkhorn} to optimize the transport plan.
Specifically, we fix text representations and class prototypes to optimize the transport plan.
The details of OT and Sinkhorn's algorithm are shown in Appendix \ref{sec:OT}.

\subsubsection{Classifying with OT Scores}
After getting the optimal transport plan, the optimal transport score between text $x_i$ and class $k$ can be computed according to the optimal transport plan and similarity matrix.
Specifically,
\begin{equation}
  \text{Dec}(x_i,k) = \sum_{m=1}^{P} \sum_{n=1}^{Q} \bm{T}^{i,k}_{m,n} \text{sim}(\mathbf{V}_{i,m},\mathbf{R}_{k,n})
\end{equation}
where Dec(,) denotes the optimal transport score between text $x_i$ and class $k$, $\bm{T}^{i,k}_{m,n}$ denotes the ($m$,$n$)-th element in the optimal transport plan between text $x_i$ and class $k$, sim(,) denotes cosine similarity, $\mathbf{V}_{i,m}$ is the text representations of $\mathcal{T}_m(x_i)$, and $\mathbf{R}_{k,n}$ is the $n$-th prototype representation of the $k$-th class.

Finally, we use the softmax function over optimal transport scores between text representations and all classes to get the predicted probability distribution, and cross-entropy loss to optimize the parameters.
Specifically,
\begin{equation}
  \mathcal{L}
  =\sum_{i=1}^{NK} \frac{-1}{NK} \text{log} \frac{\text{exp}(\text{Dec}(x_i,y_i))}{\sum_{j=1}^N \text{exp}(\text{Dec}(x_i,j))}
\end{equation}
The whole training flow is shown in Algorithm \ref{alg:train}.

\subsection{Calibrated Multi-Prompting Decoding}
We also decode the PLMs' output class scores as supplementary for the hidden states decoding to provide additional prior knowledge of the PLMs.
Specifically, we first expand the set of label words for the verbalizer, then calibrate the class scores for each prompt separately, and average the calibrated class scores of all prompts. 
Finally, we combine OT scores and calibrated class scores for joint decoding.

\subsubsection{Label Words Expansion}
We first use a simple and effective way to expand the label words.
Given the prediction layer of MLM $\mathbf{W}_{mlm} \in \mathbb{R}^{|\mathcal{V}| * d}$, we can treat each item as a representation of a word.
Then, given a label word, we expand a new set of label words $\emph{S}$ based on cosine similarity between words and use the softmax function over cosine similarity to assign a weight to each element in the set.
It is worth noting that due to the large number of open-sourced PLMs, getting an expanded set of label words is easy.

\begin{figure}[t]
\centering
\includegraphics[width=1\columnwidth]{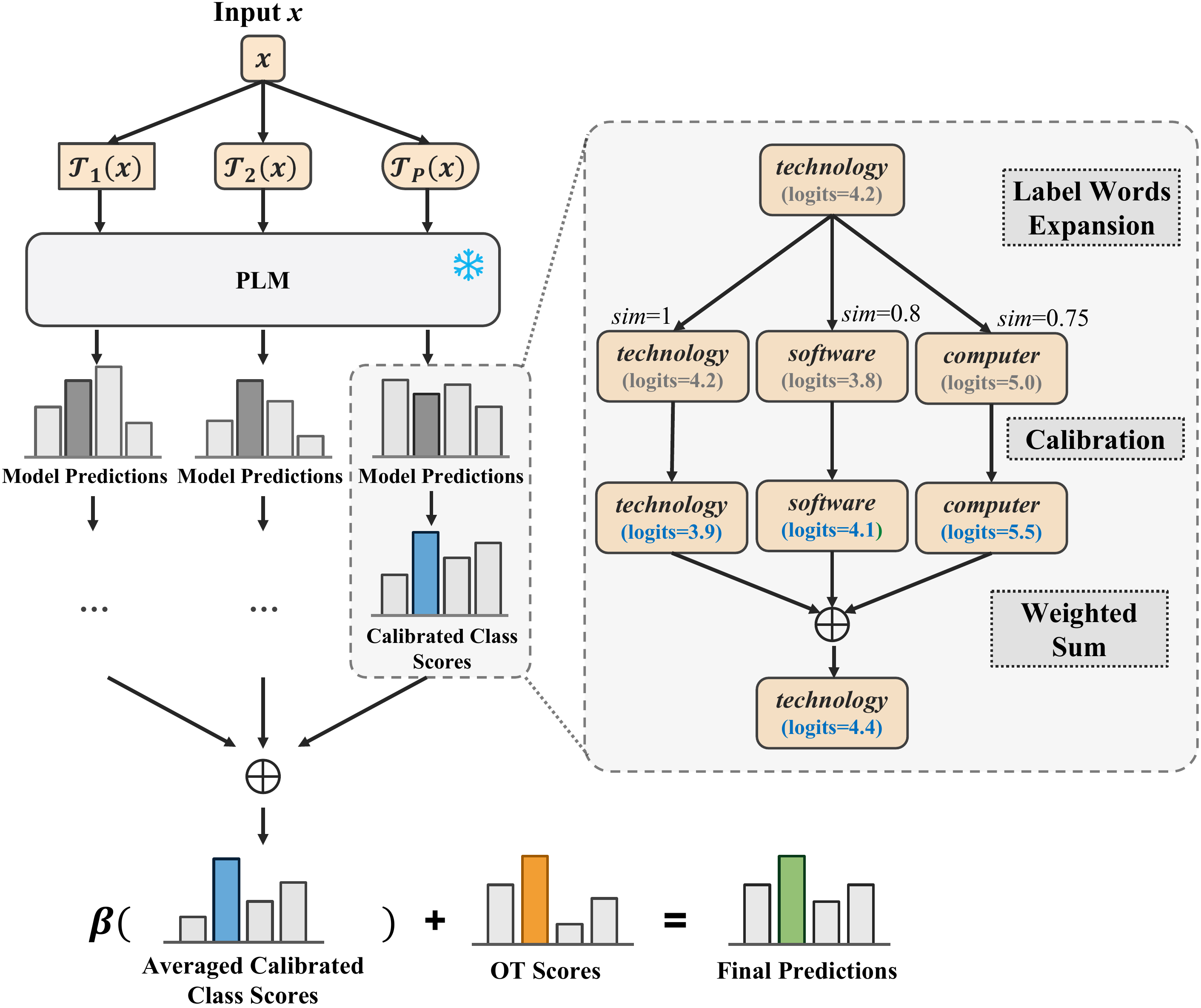}
\caption{Illustration of the calibrated multi-prompting decoding.} \label{fig:framework2}
\end{figure}

\subsubsection{Calibrating Class Scores}
Predictions of PLM are biased because PLM tends to predict tokens that are common in its pre-training distribution~\cite{DBLP:conf/icml/ZhaoWFK021}.
Thus, we first calibrate PLM on a given prompt and then use calibrated class scores for classification.

Specifically, we use a template to wrap to empty input $x_e=$``'' and query the model with $\mathcal{T}_j(x_e)$ to obtain the predictions of label words $\mathbf{s}_{ej} \in \mathbb{R}^{N|\emph{S}|}$ to calibrate.
Then, given a prediction $\mathbf{s}_{ij}$ for sample $x_i$ wrapped by prompt $\mathcal{T}_j$, we can calibrate $\mathbf{s}_{ij}$ by
\begin{equation}
    \mathbf{\bar{s}}_{ij} = \text{diag}(\mathbf{s}_{ej}/\text{mean}(\mathbf{s}_{ej}))^{-1}\mathbf{s}_{ij}
\end{equation}
where $\mathbf{\bar{s}}_{ij} \in \mathbb{R}^{N|\emph{S}|}$ is calibrated prediction over the expanded set.
We can see that the calibrated predictions for context-independent inputs such as empty input become uniformly distributed.

Then, we use the weight vector for each class to aggregate the predictions of each element in the corresponding set to form the calibrated class scores $\mathbf{\hat{s}}_{ij} \in \mathbb{R}^N$.
Finally, we average the class scores of multiple prompts to get averaged calibrated class scores $\mathbf{\hat{s}}_{i}$ for classification.
\begin{equation}
    \mathbf{\hat{s}}_{i} = \frac{\sum_{j=1}^P \mathbf{\hat{s}}_{ij}} {P}
\end{equation}

\subsubsection{Joint Decoding}
Finally, we sum OT scores and averaged calibrated class scores to jointly decode the predictions.
\begin{equation} 
 \hat{y}_{i} = \hat{y}^{OT}_i + \beta \mathbf{\hat{s}}_{i}
\end{equation}
where $\hat{y}_{i}$ is the final predictions, $\hat{y}^{OT}_i$ = [\text{Dec}($x_i$,1), $\cdots$, \text{Dec}($x_i$,$N$)] $\in \mathbb{R}^{N}$ denotes OT scores, and $\beta$ is a hyper-parameter to balance model and prior knowledge inside PLM.

\begin{table*}[t]
\resizebox{\textwidth}{!}{%
\begin{tabular}{@{}c|lccccccccccc@{}}
\toprule
\textbf{$N$} &
  \multicolumn{1}{l}{\textbf{Method}} &
  \multicolumn{1}{c}{\textbf{SST2}} &
  \multicolumn{1}{c}{\textbf{IMDB}} &
  \multicolumn{1}{c}{\textbf{Yelp}} &
  \multicolumn{1}{c}{\textbf{AG}} &
  \multicolumn{1}{c}{\textbf{DB}} &
  \multicolumn{1}{c}{\textbf{Yahoo}} &
  \multicolumn{1}{c}{\textbf{RTE}} &
  \multicolumn{1}{c}{\textbf{SNLI}} &
  \multicolumn{1}{c}{\textbf{MNLI-m/mm}} &
  \multicolumn{1}{c}{\textbf{Avg.}} \\
\midrule
\multirow{7}{*}{1} &
  \textbf{ICL} &
  $81.5_{3.7}$ &
  $65.6_{11.4}$ &
  $81.1_{10.6}$ &
  $66.7_{4.8}$ &
  $71.7_{2.6}$ &
  $53.2_{6.2}$ &
  $45.0_{4.7}$ &
  $46.1_{5.3}$ &
  ${\bf 53.6_{0.5}}/53.9_{0.8}$ &
  $61.8_{5.1}$ \\
 &
  \textbf{BBT} &
  $83.4_{1.3}$ &
  $89.0_{0.1}$ &
  $89.7_{0.1}$ &
  $75.4_{0.8}$ &
  $59.1_{1.7}$ &
  $31.2_{2.7}$ &
  $52.3_{1.4}$ &
  $38.5_{0.8}$ &
  $43.4_{2.5}/42.9_{3.3}$ &
  $60.5_{1.5}$ \\
 &
  \textbf{BBTv2} &
  $83.3_{2.5}$ &
  $89.0_{0.2}$ &
  $89.9_{0.2}$ &
  $74.3_{3.2}$ &
  $74.2_{5.2}$ &
  $34.0_{3.5}$ &
  $48.2_{5.7}$ &
  $38.6_{4.0}$ &
  $44.2_{3.2}/44.3_{4.5}$ &
  $62.0_{3.2}$ \\
 &
  \textbf{RLPrompt} &
  $63.5_{6.3}$ &
  $65.0_{6.5}$ &
  $66.3_{6.9}$ &
  $72.5_{4.5}$ &
  $65.6_{5.5}$ &
  $38.1_{5.8}$ &
  $53.8_{5.3}$ &
  $36.5_{3.0}$ &
  $40.3_{2.0}/41.0_{2.1}$ &
  $54.3_{4.8}$ \\
 &
 \textbf{PromptBoosting} &
  $86.7_{2.6}$ &
  $82.4_{6.1}$ &
  $88.7_{2.5}$ &
  $58.7_{11.8}$ &
  $73.0_{4.8}$ &
  $23.7_{7.0}$ &
  $50.0_{5.9}$ &
  $43.5_{6.1}$ &
  $36.8_{1.6}/36.3_{2.3}$ &
  $58.0_{5.1}$ \\
 &
  \textbf{DecT} &
  $90.8_{0.2}$ &
  $91.2_{0.3}$ &
  $94.8_{0.1}$ &
  $ 79.9_{1.1}$ &
  $ 78.8_{0.9}$ &
  $ {\bf 55.2_{0.8}}$ &
  $ 56.0_{2.7}$ &
  $ {\bf 47.7_{4.1}}$ &
  $52.2_{2.7}/53.3_{3.0}$ &
  $ 70.0_{1.6}$ \\
  &
  \textbf{MPD} & 
  ${\bf 92.3_{0}}$ &
  ${\bf 94.4_{0.3}}$ &
  ${\bf 95.6_{1.1}}$ &
  ${\bf 83.2_{1.5}}$&
  ${\bf 84.4_{3.5}}$&
  ${ 53.6_{0.9}}$&
  ${\bf 57.6_{1.7}}$&
    ${46.6_{4.1}}$&
  ${53.1_{1.5}/\bf54.1_{1.8}}$ &
  ${\bf 71.5_{1.6}}$
  \\
\midrule
\multirow{7}{*}{4} &
  \textbf{ICL} &
  $60.3_{9.8}$ &
  $80.4_{6.6}$ &
  $77.4_{14.6}$ &
  $65.1_{5.4}$ &
  $71.7_{6.5}$ &
  $49.9_{9.9}$ &
  $42.7_{3.9}$ &
  $42.1_{3.2}$ &
  $44.7_{5.9}/45.2_{6.0}$ &
  $58.0_{7.2}$ \\
 &
  \textbf{BBT} &
  $84.5_{1.2}$ &
  $89.8_{0.9}$ &
  $90.2_{0.6}$ &
  $79.0_{2.1}$ &
  $67.7_{3.5}$ &
  $42.9_{0.6}$ &
  $48.4_{4.0}$ &
  $40.5_{1.3}$ &
  $41.2_{1.7}/40.7_{2.0}$ &
  $62.5_{1.8}$ \\
 &
  \textbf{BBTv2} &
  $86.6_{2.2}$ &
  $89.4_{0.6}$ &
  $90.3_{0.5}$ &
  $79.1_{2.1}$ &
  $89.0_{1.7}$ &
  $46.0_{1.4}$ &
  $46.2_{2.3}$ &
  $40.8_{4.3}$ &
  $44.0_{0.9}/44.8_{1.6}$ &
  $65.6_{1.8}$ \\
 &
  \textbf{RLPrompt} &
  $80.7_{7.5}$ &
  $75.8_{10.1}$ &
  $78.8_{7.3}$ &
  $76.1_{4.8}$ &
  $76.3_{5.9}$ &
  $45.0_{3.1}$ &
  $53.5_{2.9}$ &
  $36.3_{2.6}$ &
  $44.4_{2.9}/45.5_{3.8}$ &
  $61.2_{5.1}$ \\
 &
 \textbf{PromptBoosting} &
  $88.9_{2.3}$ &
  $83.0_{5.2}$ &
  $92.3_{2.1}$ &
  $78.2_{6.8}$ &
  $90.1_{0.7}$ &
  $36.4_{5.1}$ &
  $53.5_{5.9}$ &
  $53.4_{3.4}$ &
  $39.8_{4.5}/40.3_{5.7}$ &
  $65.6_{4.2}$ \\
 &
  \textbf{DecT} &
  $87.6_{1.6}$ &
  $89.6_{0.9}$ &
  $ 94.8_{0.7}$ &
  $ 81.9_{2.6}$ &
  $89.1_{0.6}$ &
  $ 59.9_{2.1}$ &
  $ 56.7_{2.7}$ &
  $53.2_{2.9}$ &
  $ 52.2_{2.3}/53.4_{2.4}$ &
  $ 71.8_{1.9}$ \\
  &
  \textbf{MPD} & 
  ${\bf 92.6_{0.1}}$ &
  ${\bf 94.9_{0.2}}$ &
  ${\bf 96.3_{0.3}}$ &
  ${\bf 85.9_{1.0}}$ &
  ${\bf 92.8_{3.0}}$&
  ${\bf 62.2_{1.3}}$&
  ${\bf 59.2_{3.6}}$ &
  ${\bf 57.1_{2.4}}$&
  ${\bf 54.4_{0.1}/56.5_{0.6}}$ &
  ${\bf 75.2_{1.1}}$
  \\
\midrule
\multirow{7}{*}{16} &
  \textbf{ICL} &
  $71.5_{15.8}$ &
  $80.6_{6.0}$ &
  $73.7_{14.5}$ &
  $64.4_{6.0}$ &
  $71.8_{9.1}$ &
  $52.6_{5.7}$ &
  $43.8_{7.0}$ &
  $42.0_{6.3}$ &
  $51.4_{3.0}/52.1_{3.3}$ &
  $60.0_{7.7}$ \\
 &
  \textbf{BBT} &
  $89.6_{0.3}$ &
  $89.3_{0.4}$ &
  $91.5_{0.2}$ &
  $81.5_{0.8}$ &
  $87.8_{3.0}$ &
  $48.3_{1.4}$ &
  $52.6_{2.2}$ &
  $46.6_{1.3}$ &
  $40.0_{2.6}/39.9_{2.9}$ &
  $66.7_{1.5}$ \\
 &
  \textbf{BBTv2} &
  $90.3_{1.7}$ &
  $88.6_{2.1}$ &
  $92.9_{0.6}$ &
  $85.3_{0.5}$ &
  $93.6_{0.7}$ &
  $52.0_{1.4}$ &
  $56.7_{3.3}$ &
  $57.3_{2.3}$ &
  $50.1_{2.4}/51.7_{3.2}$ &
  $71.9_{1.8}$ \\
 &
  \textbf{RLPrompt} &
  $87.0_{2.6}$ &
  $87.6_{2.4}$ &
  $95.1_{1.0}$ &
  $80.2_{0.7}$ &
  $80.8_{3.3}$ &
  $48.1_{2.2}$ &
  $54.3_{2.8}$ &
  $41.1_{5.0}$ &
  $43.3_{3.9}/44.3_{4.5}$ &
  $66.2_{2.8}$ \\
 &
  \textbf{PromptBoosting} &
  $87.6_{3.0}$ &
  $86.2_{3.1}$ &
  $94.7_{1.0}$ &
  $85.2_{0.9}$ &
  $95.0_{0.5}$ &
  $46.6_{2.4}$ &
  $60.0_{5.5}$ &
  $61.3_{3.5}$ &
  $52.5_{1.5}/50.4_{5.1}$ &
  $72.0_{2.7}$ \\
 &
  \textbf{DecT} &
  $91.0_{0.5}$ &
  $ 91.0_{0.9}$ &
  $ 95.4_{0.3}$ &
  $ 86.4_{0.4}$ &
  $94.6_{0.5}$ &
  $ 64.2_{0.7}$ &
  $59.7_{1.8}$ &
  $60.5_{0.8}$ &
  $ 55.3_{1.3}/56.8_{1.5}$ &
  $ 75.5_{0.9}$ \\
  &
  \textbf{MPD} & 
  ${\bf 91.9_{0.1}}$ &
  ${\bf 95.1_{0.1}}$ &
  ${\bf 96.4_{0.1}}$&
   ${\bf 87.9_{0.4}}$ &
   ${\bf 96.7_{0.2}}$ &
   ${\bf 68.3_{0.3}}$ &
   ${\bf 61.7_{2.8}}$&
   ${\bf 62.4_{1.3}}$&
   ${\bf 57.5_{1.1}/59.7_{1.1} }$ &
   ${\bf 77.8_{0.8}}$
  \\
\bottomrule
\end{tabular}%
}
\caption{Experiment results for MaaS adaptation methods. The results (i.e., average accuracy and standard deviation (\%)) over 5 runs. The best results are in \textbf{bold}.} \label{tab:results}
\end{table*}

\section{Experiments}

\subsection{Datasets}
We conduct experiments on several common natural language understanding tasks including sentiment analysis, topic classification, and natural language inference (NLI).
For sentiment analysis, we choose SST2~\cite{DBLP:conf/emnlp/SocherPWCMNP13}, IMDB~\cite{DBLP:conf/acl/MaasDPHNP11}, and Yelp~\cite{DBLP:conf/nips/ZhangZL15}.
For topic classification, we choose AG's News~\cite{DBLP:conf/nips/ZhangZL15}, DBPedia~\cite{DBLP:conf/nips/ZhangZL15}, and Yahoo~\cite{DBLP:conf/nips/ZhangZL15}. 
For NLI, we choose RTE~\cite{DBLP:conf/mlcw/DaganGM05}, SNLI~\cite{DBLP:conf/emnlp/BowmanAPM15}, and MNLI~\cite{DBLP:conf/naacl/WilliamsNB18}.
The statistics of datasets are shown in Table \ref{tab_dataset}.

We also randomly sample $N$ = 1, 4, 16 samples for each class from the original training set to construct the few-shot training set.
Following~\citet{DBLP:conf/acl/CuiLDHLS23} and~\citet{hou2023promptboosting}, we use the original validation set for evaluation on the GLUE (i.e., SST, RTE, and MNLI) and SNLI datasets.

\subsection{Baselines}
We compare with some strong MaaS methods.
\textbf{In-context learning (ICL)}~\cite{DBLP:conf/nips/BrownMRSKDNSSAA20} concatenates some text-label pairs before the test samples.
\textbf{BBT}~\cite{DBLP:conf/icml/SunSQHQ22} optimizes soft prompt tokens with a derivative-free evolutionary algorithm.
\textbf{BBTv2}~\cite{DBLP:conf/emnlp/SunHQZHQ22} further prepends continuous prompts to every layer and uses a divide-and-conquer algorithm to optimize them.
\textbf{RLPrompt}~\cite{DBLP:conf/emnlp/DengWHWGSSXH22} optimizes discrete prompts with RL and adopts soft Q-learning to find the optimal prompt.
\textbf{PromptBoosting}~\cite{hou2023promptboosting} sequentially learns multiple weak learners and adopts AdaBoost to ensemble them.
\textbf{DecT}~\cite{DBLP:conf/acl/CuiLDHLS23} extracts text feature and optimizes class prototypes as hyperspheres with an additional radius parameter.

\subsection{Implementation Details}
For a fair comparison with baselines, we also use  RoBERTa$_{\texttt{LARGE}}$~\cite{DBLP:journals/corr/abs-1907-11692liu} as the backbone model.
We set the text representation dimension to 128 and optimize the parameters for 30 epochs with Adam optimizer~\cite{kingma2014adam}.
We set the number of templates $P$ to 3 and the number of prototypes $Q$ to 3.
We set $\lambda$ to 0.1 and threshold 0.01 in the Sinkhorn's algorithm.
Following \citet{DBLP:conf/acl/CuiLDHLS23}, we directly set $\beta=1/n$ for most datasets based on the intuition that $\beta$ should decrease as the amount of training data increases.
We set $\beta=1$ on the MNLI dataset based on the validation set.
We expand 10 label words for each class and we do not use label words expansion on the NLI datasets.
This is because NLI datasets have a smaller number of potential or related label words that express the class compared to other datasets (i.e., sentiment analysis and topic classification).
We give the templates and label words in Table \ref{tab:template}.

\begin{table}[t]
\resizebox{\linewidth}{!}{%
\begin{tabular}{l|cc|cc|cc}
\toprule
\multirow{2}{*}{\textbf{Method}} & \multirow{2}{*}{\makecell[c]{\textbf{Trainable}\\\textbf{Param (K)}}} & \multirow{2}{*}{\makecell[c]{\textbf{Query}\\\textbf{Number}}} & \multicolumn{2}{c|}{\textbf{SST2}} & \multicolumn{2}{c}{\textbf{AG}}  \\
         & & & \textbf{Acc} & \textbf{Training Time (s)}  & \textbf{Acc} & \textbf{Training Time (s)} \\
         \midrule
\textbf{ICL}& 0   & 0   & 71.5  & 0 & 64.4 & 0  \\
\textbf{BBT} & 0.5   & 8,000  &89.6 & 1619 & 87.8 & 5228\\
\textbf{BBTv2}& 12  & 8,000  & 90.3 & 1624& 85.3 &  6546                    \\
\textbf{RLPrompt} & 3100  & 12,000   &87.0 & 82286 & 80.2 & 99947        \\
\textbf{PromptBoosting} & 0.4 & 10 & 87.6 & 190 &85.2 &398\\
\textbf{DecT}     & 130     & 1    & 91.0 &1.4 & 86.4 & 2.3\\
\textbf{Ours} & 132 & 3 & 91.9  & 3.5& 87.9 & 6.7 \\ 
\bottomrule
\end{tabular}%
}
\caption{Efficiency comparison for MaaS methods under 16-shot setting on two datasets.} \label{tab:eff}
\end{table}

\subsection{Results}
Table \ref{tab:results} shows the performance on 9 datasets under the few-shot setting.
We can see that our model achieves state-of-the-art performance.
Specifically, compared to DecT, our approach achieves 1.5\%, 3.4\%, and 2.3\% average performance improvement under 1,4,16-shot settings.
Among all settings, our model performs the best in 27 out of the 30.
This shows our model is effective and the improvement is consistent.
It is worth noting that our model performs well on both simple (i.e., sentiment analysis and topic classification) and difficult tasks that require semantic understanding (i.e., NLI).
This shows that the improvement is task-independent.

By observing other baseline methods, ICL does not achieve performance gains, as the shot number increases.
This is due to the length limitation of the PLM.
BBTv2 exceeds BBT and RLPrompt.
This shows that inserting continuous prompts to every layer of the PLM is effective.
PromptBoosting and DecT usually achieve better performance than BBT and BBTv2.
This is because these two methods do not need to search prompts in a large space and can benefit from gradients.

In addition, our method also achieves the minimum average standard deviation across different datasets and settings.
This shows that our approach is the most robust.
As the shot number increases, our model achieves the minimum standard deviation on 1, 6, and 8 datasets.
This shows that compared to other methods, our model is more likely to obtain a stable result as the shot number increases.

\begin{table*}[t]
\resizebox{\textwidth}{!}{%
\begin{tabular}{@{}c|lccccccccccc@{}}
\toprule
\textbf{$N$} &
  \multicolumn{1}{l}{\textbf{Method}} &
  \multicolumn{1}{c}{\textbf{SST2}} &
  \multicolumn{1}{c}{\textbf{IMDB}} &
  \multicolumn{1}{c}{\textbf{Yelp}} &
  \multicolumn{1}{c}{\textbf{AG}} &
  \multicolumn{1}{c}{\textbf{DB}} &
  \multicolumn{1}{c}{\textbf{Yahoo}} &
  \multicolumn{1}{c}{\textbf{RTE}} &
  \multicolumn{1}{c}{\textbf{SNLI}} &
  \multicolumn{1}{c}{\textbf{MNLI-m/mm}} &
  \multicolumn{1}{c}{\textbf{Avg.}} \\
\midrule
\multirow{3}{*}{64} &
  \textbf{Fine-tuning} &
  $92.5_{1.9}$ &
  $86.3_{3.8}$ &
  $94.5_{1.4}$ &
  $87.4_{0.6}$ &
  $98.2_{0.2}$ &
  $69.0_{0.7}$ &
  $67.7_{3.2}$ &
  $66.6_{6.4}$ &
  $65.6_{2.9}/67.7_{4.0}$ &
  $79.6_{2.5}$ \\
 &
  \textbf{DecT} &
  $92.4_{0.5}$ &
  $91.3_{0.5}$ &
  $94.9_{0.5}$ &
  $89.2_{0.3}$ &
  $97.0_{0.1}$ &
  $69.3_{0.4}$ &
  $65.7_{1.7}$ &
  $67.2_{1.0}$ &
  $62.0_{1.4}/63.3_{1.3}$ &
  $79.2_{0.8}$ \\
   &
  \textbf{MPD} &
  $92.0_{0.2}$ &
  $95.1_{0.5}$ &
  $96.6_{0.1}$ &
  $89.6_{0.3}$ &
  $97.4_{0.1}$ &
  $70.6_{0.3}$ &
  $67.6_{1.2}$ &
  $68.2_{0.7}$ &
  $63.4_{0.7}/65.1_{1.1}$ &
  $80.6_{0.5}$ \\
\midrule
\multirow{3}{*}{256} &
  \textbf{Fine-tuning} &
  $92.0_{0.9}$ &
  $92.1_{0.2}$ &
  $94.3_{0.3}$ &
  $89.6_{0.3}$ &
  $98.5_{0.2}$ &
  $70.2_{0.4}$ &
  $79.8_{1.0}$ &
  $84.4_{0.4}$ &
  $77.2_{0.2}/78.7_{0.3}$ &
  $85.7_{0.4}$ \\
 &
  \textbf{DecT} &
  $92.7_{0.2}$ &
  $92.1_{0.1}$ &
  $95.6_{0.1}$ &
  $90.3_{0.1}$ &
  $97.4_{0.1}$ &
  $71.3_{0.1}$ &
  $69.2_{1.0}$ &
  $69.7_{0.4}$ &
  $68.0_{0.3}/69.4_{0.3}$ &
  $81.6_{0.3}$ \\
   &
  \textbf{MPD} &
  $92.3_{0.2}$ &
  $95.1_{0.1}$ &
  $96.7_{0}$ &
  $90.6_{0.2}$ &
  $97.6_{0.1}$ &
  $71.4_{0.2}$ &
  $71.6_{1.0}$ &
  $70.3_{0.6}$ &
  $68.9_{0.4}/69.9_{0.7}$ &
  $82.4_{0.4}$ \\

\bottomrule

\end{tabular}%
}
\caption{Experiment results for more training data. The results over 5 runs. Fine-tuning method requires gradients of the model to update the model.
} \label{tab:ft}
\end{table*}

\begin{table}\small  \setlength{\tabcolsep}{12pt}
\centering
\begin{tabular}{l|cccc}
\toprule
\multirow{2}{*}{\textbf{Model Setting}}& \multicolumn{3}{c}{\textbf{Average Accuracy}} \\
    \cmidrule{2-4}
      & 1 & 4 & 16 \\
\midrule
\textbf{Ours}& \textbf{71.5} & \textbf{75.2} & \textbf{77.8}\\
\midrule
\textbf{w/o Class Scores} & 57.8 & 69.6 & 76.5\\
\textbf{w/o OT Scores} & 68.0  & 68.0 & 68.0  \\
\textbf{w/o Expansion}& 70.7 & 74.8 & 77.7  \\
\bottomrule
\end{tabular}
\caption{The ablation study of our model on all datasets. We run each experiment over 5 random seeds.}  \label{tab:ablation}
\end{table}

\subsection{Efficiency Comparison}
We further compare our model with baselines in terms of efficiency, including the number of trainable parameters, the number of queries to access PLM, and training time.
As shown in Table~\ref{tab:eff}, our model is effective and efficient.
The efficiency of our approach significantly outperforms all baselines except DecT.
Specifically, our model only queries PLM three times per training data, which is about 3$\times$ (2500$\times$) less than PromptBoosting (BBT).
In addition, the training time of our method is about 50$\times$ (400$\times$) faster than that of PromptBoosting (BBT).
Our method uses multiple prompts, so the training time is slightly longer than in DecT.
Overall, the training speed of our method is still fast.
In addition, our model only includes 132K trainable parameters, significantly smaller than the number of parameters in RoBERTa$_{\texttt{LARGE}}$ (i.e., about 0.04\%).
Even though some methods (i.e., BBT, BBTv2, and PromptBoosting) have fewer trainable parameters, our model takes less time to train and achieves better performance.

\subsection{Performance on More Training Data}
We conduct experiments with a larger training set (i.e., $N =$ 64 and 256) to explore the scalability of our model beyond the few-shot setting.
We compare our model with DecT and a strong baseline fine-tuning.
Fine-tuning requires gradients of PLM to update full parameters of PLM.

As shown in Table \ref{tab:ft}, firstly, our model and DecT both achieve performance improvement when the shot number increases.
This shows that these two methods can still fit larger datasets even with a small number of tunable parameters.
In addition, as the shot number increases, the average standard deviation of two models decreases.
Secondly, our model also exceeds DecT when $N$ is large.
This shows that the improvement is consistent.
It is worth noting that the gap between the two methods is narrowing as the shot number increases.
Thirdly, our method exceeds fine-tuning in the 64-shot setting and is exceeded by fine-tuning in the 256-shot setting.
This is because our method cannot update PLM parameters, it can only update a small portion of parameters.
Thus, when the shot number is larger, our model may not be able to learn more knowledge.
Finally, similar to DecT, our model does not show significant performance differences compared to fine-tuning in simple tasks (i.e., sentiment analysis and topic classification).
However, there are significant differences in harder tasks (i.e., SNLI).
This suggests that harder tasks require more parameters to learn corresponding semantics.

\section{Model Analysis}

\subsection{Ablation Study}
We conduct ablation study to validate the effectiveness of each component.
As shown in Table \ref{tab:ablation}, firstly, all components achieve performance improvement.
Secondly, the performance of w/o Class Scores (i.e., only using OT scores) improves rapidly as the shot number increases.
This shows that shot number has a significant effect on OT scores.
Thirdly, w/o OT Scores (i.e., only averaging calibrated class scores of multiple prompts) also achieves good performance, especially in the 1-shot setting.
This shows that averaging the class scores of multiple prompts is also a strong baseline under 1-shot setting.
Finally, label words expansion achieves more significant performance gains when the shot number is small.
This is because the model relies more on class scores when the shot number is small.
The results of label words expansion are shown in Appendix \ref{sec:expansion}.

\begin{figure*}[t]
\centering
\includegraphics[width=1.8\columnwidth]{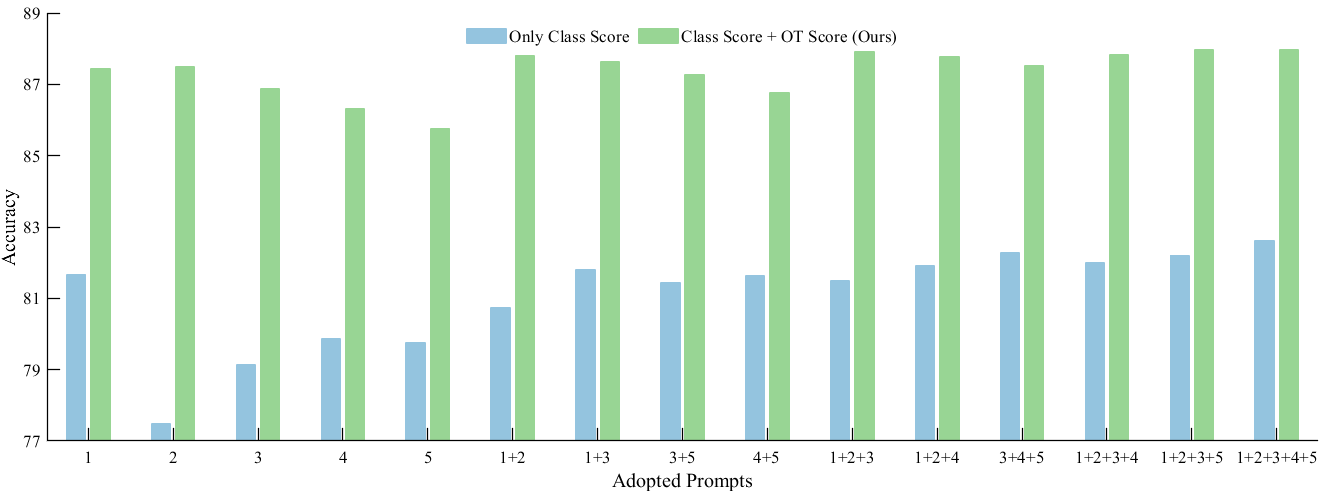}
\caption{Effects of prompts on AG's news dataset under 16-shot setting. 1, 2, 3, 4, and 5 represent 5 different prompts. 1+2 denotes we use two prompts (i.e., 1 and 2) to train the model. The results over 5 runs.} \label{fig:tempalte}
\end{figure*}

\subsection{Effects of Prompts}
We explore the effects of prompts by changing the combination of prompts in Figure \ref{fig:tempalte}.
Specifically, we use different prompts and change the number of prompts.
The five templates are shown in Table \ref{tab:effects_of_prompt}.

Firstly, as the prompt number increases, performance will be better and more stable.
When the number of prompts exceeds 3, the improvement is marginal.
Secondly, there will be less performance fluctuation with multiple prompts compared to using a single prompt
Specifically, when we use one/two/three/four prompts, the change in performance is 1.7\%/0.3\%/0.2\%/0.2\%.
This shows that our method is robust across multiple prompts.
Thirdly, the good performance of a prompt's class scores does not mean that using it can achieve better performance.
This shows that there is a gap between the representation of hidden states and class scores.

\begin{figure}[t]
\centering
\includegraphics[width=1\columnwidth]{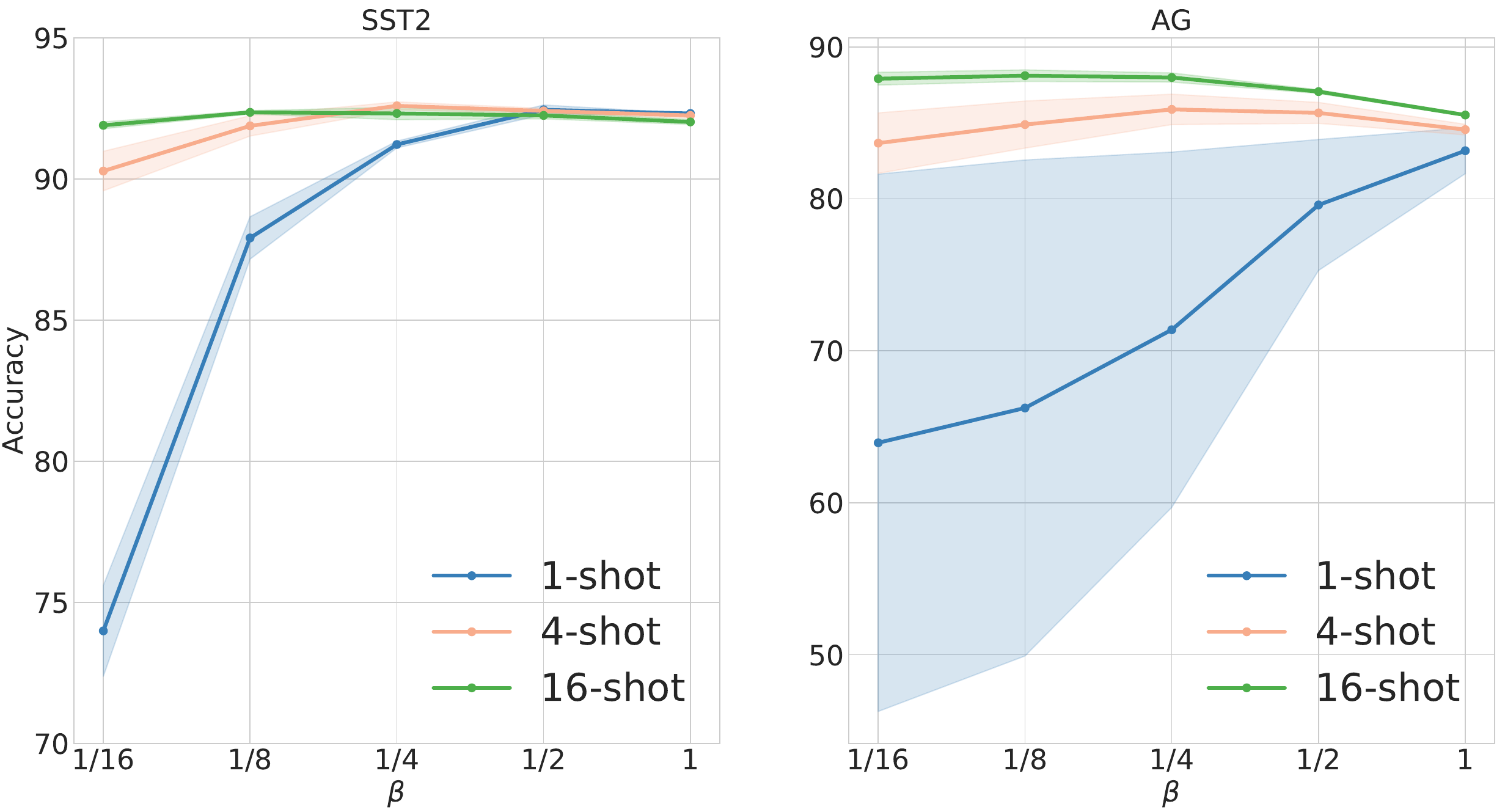}
\caption{Effects of $\beta$ over 5 runs on two datasets.} \label{fig:beta}
\end{figure}

\subsection{Effects of $\beta$}
Hyper-parameter $\beta$ controls the ratio of OT scores and class scores.
We explore the effects of $\beta$ on two datasets in Figure \ref{fig:beta}.
Firstly, our model needs a bigger $\beta$ under 1-shot setting.
This is because the amount of data is small and the model relies heavily on class scores for prediction rather than OT scores.
Secondly, as the shot number increases, the effect of $\beta$ on the results diminishes.
This is because as the shot number increases, the model's capability is enhanced and relies more on the OT scores for prediction.
Thirdly, as the shot number increases, the standard deviation becomes smaller.
This shows that the OT scores are more stable when the training size increases.

\begin{figure}[t]
\centering
\includegraphics[width=1\columnwidth]{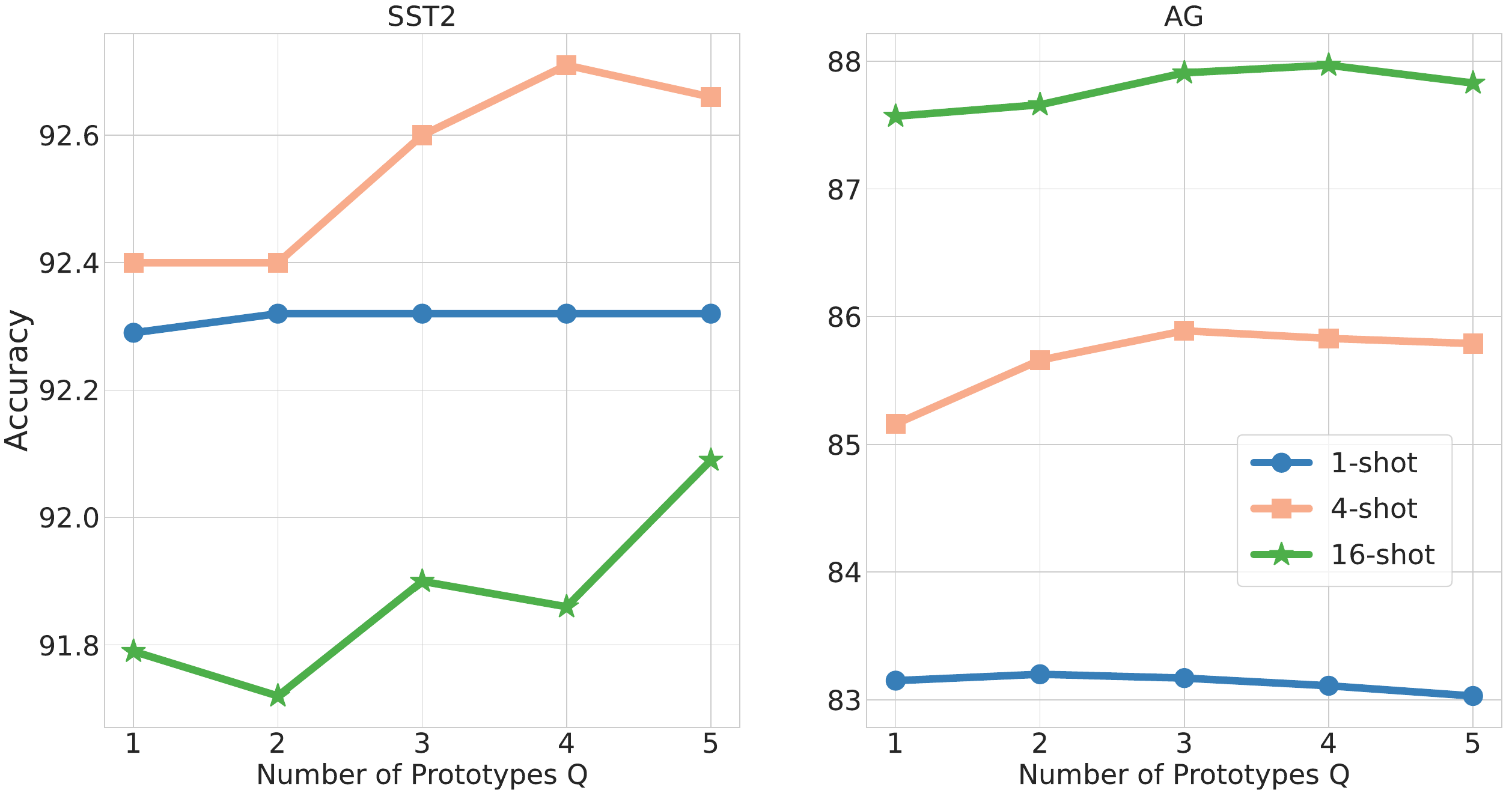}
\caption{Effects of $Q$ over 5 runs on two datasets.} \label{fig:Q}
\end{figure}

\vspace{-0.2em}
\subsection{Effects of the Number of Prototypes}
We explore the effects of the number of prototypes on two datasets in Figure \ref{fig:Q}.
We can see that using multiple prototypes tends to outperform using one.
This shows that using multiple prototypes is valid in our multiple prompts setting.
As the shot number increases, a large $Q$ usually performs better.
This is because when the shot number increases, the model needs a large $Q$ to capture subtle differences between multiple texts of the same class.

\vspace{-0.3em}
\section{Conclusion}
In this paper, we propose the multi-prompting decoder framework for output-side MaaS adaptation on few-shot tasks. 
To achieve better decoding performance, we design two decoding strategies to decode PLMs' output hidden states and class scores, respectively. 
Extensive experiments show that our approach is effective and efficient.

\section*{Limitations}
Although our approach achieves state-of-the-art performance, it has some limitations. 

The first limitation is that our approach relies on two kinds of PLMs' outputs, namely hidden states and class scores, but these outputs are not always available for some PLMs, such as ChatGPT.
More rigorous decoding, such as decoding only based on the accessing of the output tokens, is worth further exploration.

The second limitation is that we have only conducted experiments on the English datasets.
The performance of our method on datasets in other languages deserves further exploration.

The third limitation is that our method relies on some manual engineering, such as the design of prompts and label words. 
In this paper, this problem is alleviated by using templates and label words from previous works.

The fourth limitation is that MPD may not be suitable for training with more training data.
As shown in Table \ref{tab:results} and Table \ref{tab:ft}, the performance gap of between MPD and DecT is narrowing as the shot number $N$ increases from 4 to 256.
When $N$ is 4, 16, 64, and 256, the average performance gap is 3.4, 2,3, 1.4, and 0.8, respectively.
Considering that as the shot number $N$ increases, the performance improvement becomes marginal, and using multiple prompts for querying incurs additional costs, we think that our MPD is not suitable in this situation.

\section*{Acknowledgments}
We would like to thank the anonymous reviewers for their insightful comments.
This work is supported by the National Natural Science Foundation of China under Grants Nos. 62441225, 61972192, 62172208, 61906085.
This work is partially supported by Collaborative Innovation Center of Novel Software Technology and Industrialization.
This work is supported by the Fundamental Research Funds for the Central Universities under Grant No. 14380001.

\bibliography{anthology,custom}

\appendix
\clearpage

\section{Optimal Transport} \label{sec:OT}

OT aims to find the minimal cost transport plan between two distributions.
In this paper, we focus on discrete distribution and define two discrete distributions as follows:
\begin{equation} \label{eq:discrete}
U=\sum_{p=1}^{P}u_p\delta_{\bm{f}_p} \hspace{2em} \text{and} \hspace{2em} V=\sum_{q=1}^{Q}v_q\delta_{\bm{g}_q},
\end{equation}
where $\bm{u}$ and $\bm{v}$ are the discrete probability vectors that sum to 1, and $\delta_{\bm{f}}$ is a Dirac delta function operating on $\bm{f}$ in the embedding space.
Then, the cost matrix $\bm{C} \in \mathbb{R}^{P \times Q}$ in OT can be defined as $\bm{C}_{p,q}= 1- \text{sim}(\bm{f}_p,\bm{g}_{q})$ and each element in the cost matrix denotes the cost between $\bm{f}_p$ and $\bm{g}_q$, where sim(,) denotes a similarity metric function.
The optimization problem of optimal transport is formulated as:
\begin{equation} \label{eq:optimization}
\begin{aligned}
& \underset{\bm{T}}{\text{minimize}}  \quad  \sum_{p=1}^{P} \sum_{q=1}^{Q} \bm{T}_{p,q} \bm{C}_{p,q}\\
& \text{subject to}\ \bm{T}\bm{1}_Q = \bm{u},\ \bm{T}^\top \bm{1}_P = \bm{v},\ \bm{T} \in \mathbb{R}^{P \times Q}_+.
\end{aligned}
\end{equation}
where $\bm{T}$ is the learned transport plan denotes the matching flow between two distributions and $\bm{T}_{p,q}$ is the amount of mass that needs to move $\bm{f}_p$ to $\bm{g}_q$.

Since direct optimization of the above processes is usually time-consuming, \citet{cuturi2013sinkhorn} use Sinkhorn distance which adds an entropic constraint for fast optimization.
Thus, the new optimization objective is formulated as:
\begin{equation} \label{eq:Sinkhorn}
\begin{aligned}
& \underset{\bm{T}}{\text{minimize}}  \quad \sum_{p=1}^{P} \sum_{q=1}^{Q} \bm{T}_{p,q} \bm{C}_{p,q} - \lambda h(\bm{T})\\
& \text{subject to}\ \bm{T}\bm{1}_Q = \bm{u},\ \bm{T}^\top \bm{1}_P = \bm{v},\ \bm{T} \in \mathbb{R}^{P \times Q}_+,
\end{aligned}
\end{equation}
where $h(\cdot) $ is entropy and $\lambda \geq 0$ is a hyper-parameter. Then we can have a fast optimization solution to get optimal transport plan $\bm{T}^*$ with a few iterations as:
\begin{equation} \label{eq:Sinkhorn_optimization}
\bm{T}^*= \text{diag}(\bm{u}^{(t)}) \exp(-\bm{C}/\lambda) \text{diag}(\bm{v}^{(t)}),
\end{equation}
where $t$ denotes the iteration and in each iteration 
$\bm{u}^{(t)} =\bm{u}/(\exp(-\bm{C}/\lambda)\bm{v}^{(t-1)})$ and $\bm{v}^{(t)} =\bm{v}/\left(\exp(-\bm{C}/\lambda)^\top\bm{u}^{(t)}\right)$, with the uniform initiation $\bm{u}$ and $\bm{v}$, and $\bm{v}^{(0)} = \bm{1}_Q$.

\section{Datasets} \label{sec:dataset}
In this section, we describe the datasets used in Table`\ref{tab:results}.

The sentiment analysis task includes three datasets.
The Stanford Sentiment Treebank (SST2) dataset and IMDB dataset were both extracted from movie reviews.
The Yelp dataset consists of reviews from Yelp.

The topic classification task includes three datasets.
AG's news dataset collects more than 1 million news articles from more than 2,000 news sources.
We also only use the title and description fields to classify news.
The Yahoo dataset was collected from the Yahoo Webscope program and contains questions and their answers.
The fields we used include question title, question content, and best answer.
The DBPedia ontology classification dataset is constructed by picking 14 non-overlapping classes from DBPedia 2014 derived from Wikipedia.

The NLI task includes three datasets.
The Recognizing Textual Entailment (RTE) dataset comes from a series of textual entailment challenges.
Examples on the RTE dataset are constructed based on news and Wikipedia text.
The Stanford Natural Language Inference (SNLI) corpus is a collection of human-written English sentence pairs manually labeled for balanced classification with the labels entailment, contradiction, and neutral.
MNLI (Multi-genre Natural Language Inference) offers ten distinct genres of written and spoken English NLI data.
MNLI-m and MNLI-mm correspond to versions matched and mismatched of the MNLI dataset, respectively.

\begin{table}[t]
\resizebox{\linewidth}{!}{%
\begin{tabular}{ccccc}
\toprule
\textbf{Task}                                  & \textbf{Dataset}        & \textbf{\# Class}  & \textbf{\# Test}   \\
\midrule
\multirow{3}{*}{\makecell[c]{Sentiment\\ Analysis}}   & SST2         & 2                  & 872       \\
 & Yelp         &2                  & 38,000     \\
 & IMDB           & 2                  & 25,000     \\ \midrule
\multirow{3}{*}{\makecell[c]{Topic\\ Classification}} & AG's News         & 4     & 7,600      \\
  & Yahoo          & 10                 & 60,000     \\
 & DBPedia        & 14                 & 70,000     \\ \midrule
\multirow{4}{*}{NLI}                  & RTE            & 2                  & 277       \\
 & SNLI           & 3                  & 9,842      \\
 & MNLI-m & 3                  & 9,815 \\  
 & MNLI-mm & 3                  & 9,832 \\  
\bottomrule
\end{tabular}
}
\caption{Statistics of datasets.}  \label{tab_dataset}
\end{table}

\begin{table*}[h]
\resizebox{\textwidth}{!}{%
\begin{tabular}{@{}llllllllllll@{}}
\toprule
\textbf{Model} &
  \multicolumn{1}{c}{\textbf{SST2}} &
  \multicolumn{1}{c}{\textbf{IMDB}} &
  \multicolumn{1}{c}{\textbf{Yelp}} &
  \multicolumn{1}{c}{\textbf{AG}} &
  \multicolumn{1}{c}{\textbf{DB}} &
  \multicolumn{1}{c}{\textbf{Yahoo}} &
  \multicolumn{1}{c}{\textbf{RTE}} &
  \multicolumn{1}{c}{\textbf{SNLI}} &
  \multicolumn{1}{c}{\textbf{MNLI-m/mm}} &
  \multicolumn{1}{c}{\textbf{Avg.}} \\
\midrule
\textbf{T5$_{\texttt{Base}}$} &
  $86.9_{0.6}$ &
  $87.5_{0.6}$ &
  $90.8_{0.4}$ &
  $85.7_{0.7}$ &
  $92.9_{0.3}$ &
  $63.9_{0.6}$ &
  $58.3_{0.4}$ &
  $60.3_{2.2}$ &
  $59.0_{1.0}/60.9_{1.1}$ &
  $74.6_{0.8}$ \\
\textbf{T5$_{\texttt{Large}}$} &
  $92.3_{0.4}$ &
  $92.6_{0.6}$ &
  $95.5_{0.1}$ &
  $86.5_{0.6}$ &
  $95.6_{0.4}$ &
  $66.4_{0.6}$ &
  $63.3_{2.7}$ &
  $61.6_{1.4}$ &
  $59.8_{2.9}/62.5_{3.0}$ &
  $77.6_{1.3}$ \\
\textbf{T5$_{\texttt{3B}}$} &
  $91.8_{0.7}$ &
  $93.0_{0.6}$ &
  $95.6_{0.1}$ &
  $87.5_{0.6}$ &
  $95.9_{0.2}$ &
  $67.6_{0.7}$ &
  $68.1_{1.0}$ &
  $64.7_{1.0}$ &
  $61.5_{1.2}/63.5_{1.5}$ &
  $78.9_{0.8}$ \\
\bottomrule
\end{tabular}%
}
\caption{Performance (16-shot) for our method on different versions of T5~\cite{raffel2020exploring}. We also run each experiment over 5 random seeds and report the average accuracy and standard deviation (\%).}  \label{tab:scaling}
\end{table*}

\section{Comparsion with TEMPERA and DP$_2$O}
We further compare our model with two existing state-of-the-art methods (i.e., TEMPERA~\cite{DBLP:conf/iclr/Zhang0ZSG23} and DP$_2$O~\cite{li2023dialogue}) following their experimental settings on four sentiment classification datasets (i.e., SST2~\cite{DBLP:conf/emnlp/SocherPWCMNP13}, MR~\cite{DBLP:conf/acl/PangL05}, CR~\cite{DBLP:conf/kdd/HuL04}, and Yelp~\cite{DBLP:conf/nips/ZhangZL15}).
They both use RL to construct or match prompts for each input.
Specifically, TEMPERA designs the agent to perform different editing techniques to construct query-dependent prompts. 
DP$_2$O first designs a prompt generation strategy based on GPT-4 and then designs a RL framework based on policy gradients to match suitable prompts for a single input.

Our results demonstrate that our model consistently outperforms these baselines in terms of both average performance and standard deviations, showcasing the effectiveness and robustness of our method.
Moreover, we observe that MPD also significantly reduces training time compared to the two methods.

\begin{table}[t] \small \setlength{\tabcolsep}{2pt}
\begin{tabular}{lccccccc}
\toprule
  \multicolumn{1}{l}{\textbf{Method}} &
  \multicolumn{1}{c}{\textbf{SST2}} &
  \multicolumn{1}{c}{\textbf{MR}} &
  \multicolumn{1}{c}{\textbf{CR}} &
  \multicolumn{1}{c}{\textbf{Yelp}} &
  \multicolumn{1}{c}{\textbf{Avg.}} \\
\midrule
   \textbf{TEMPERA}   & $91.9_{2.0}$ & $88.0_{1.1}$  & $91.1_{1.6}$ & $92.6_{1.7}$ & $90.9_{1.6}$\\
 $\bf{DP_2O}$   & $\bf{93.6_{0.7}}$ & $88.6_{0.9}$  & $90.8_{0.5}$ & $94.3_{0.4}$ & $91.8_{0.6}$ \\
 \textbf{MPD}  & $92.1_{0.1}$ & $\bf{89.6_{0.2}}$ & $\bf{91.3_{0.3}}$ & $\bf{96.4_{0.1}}$ & $\bf{92.4_{0.2}}$\\
\bottomrule
\end{tabular}%
\caption{Comparsion with TEMPERA and DP$_2$O following their experimental settings. The results (i.e., average accuracy and standard deviation (\%)) over 4 runs. The best results are in \textbf{bold}.
} \label{tab:result2}
\end{table}

\section{Effectiveness of OT}
We use two variants to replace $\bm{T}$ to illustrate the effectiveness of OT, including uniform distribution and cosine similarity.

We can see that OT achieves the best performance in Table \ref{tab:OT}.
This shows that OT can effectively match the relationship between test representations and prototypes.
The uniform distribution does not take into account the similarity between two elements.
Cosine similarity only calculates the similarity between two elements and cannot take into account the similarity with other elements.

\begin{table}[h]\small  \setlength{\tabcolsep}{10pt}
\centering
\begin{tabular}{l|cccc}
\toprule
\multirow{2}{*}{\textbf{Model Setting}}& \multicolumn{3}{c}{\textbf{Average Accuracy}} \\
    \cmidrule{2-4}
      & 1 & 4 & 16 \\
\midrule
\textbf{Ours}& \textbf{71.5} & \textbf{75.2} & \textbf{77.8}\\
\midrule
\textbf{Uniform Distribution} & 70.7 & 74.7 & 77.1\\
\textbf{Cosine Similarity} & 68.3 & 73.8  & 77.4  \\
\bottomrule
\end{tabular}
\caption{Performance of two variants to replace optimal transport plan. We run each experiment over 5 random seeds.}  \label{tab:OT}
\end{table}

\section{Visualization of Label Words Expansion} \label{sec:expansion}
We show the expanded label words in Table \ref{tab:expand}.
We set the size of expanded label words to 10.
We can see that the expansion method can find some relevant words and improve performance.
For example, on the three sentiment datasets (i.e., SST2, IMDB, and Yelp), label word `bad' expands `poor', `wrong', and `worst', and label word `great' expands `good', `perfect', and `powerful'.
While `bad' extends `good', we assign a weight to `good' based on its cosine similarity to the target class word to avoid the effect of noise.

\section{Model Scaling}
We further explore how our model applies to PLMs with different architectures and scales.

We use a encoder-decoder architecture PLM T5~\cite{raffel2020exploring} with different scales, from T5$_{\texttt{Base}}$, T5$_{\texttt{Large}}$ to T5$_{\texttt{3B}}$ in Table~\ref{tab:scaling}.
Firstly, MPD also achieves good performance on a encoder-decoder architecture PLM T5, illustrating the ability of MPD to transfer across PLM architectures.
Secondly, a larger model often achieves better performance.
This is because larger models have more representative power to get better output, including hidden states and class scores.

\section{Trade-offs between Training Time and Performance}
In this section, we further explore the trade-offs between training time and performance in our method.

Firstly, training time is proportional to the number of adopted prompts from Table~\ref{tab:trade}.
This is because we need to feed more training data into PLM.
Secondly, when the number of prompts exceeds 3, the improvement becomes marginal.
As a result, to balance training time and performance, we use three prompts.

\begin{table}[h] \setlength{\tabcolsep}{4pt}
\begin{tabular}{lccccccc}
\toprule
  \multicolumn{1}{l}{\textbf{Adopted Prompts}} &
  \multicolumn{1}{c}{\textbf{Acc}} &
  \multicolumn{1}{c}{\textbf{Training Time (s)}} \\
\midrule
   \textbf{1}   &  87.44 & 2.3\\
 \textbf{1+2}   & 87.82 & 4.5\\
 \textbf{1+2+3}   & 87.94&6.7\\
 \textbf{1+2+3+5}   & 87.98&8.8\\
 \textbf{1+2+3+4+5}   & 87.97&10.9\\
\bottomrule
\end{tabular}%
\caption{Accuracy and training time using different prompts on AG's news dataset under 16-shot setting.} \label{tab:trade}
\end{table}

\begin{table}[h] \setlength{\tabcolsep}{25pt}
\centering
\begin{tabular}{lccccccc}
\toprule
  \multicolumn{1}{l}{\textbf{Adopted Prompts}} &
  \multicolumn{1}{c}{\textbf{Acc}} \\
\midrule
\textbf{Vani}   &  87.44 \\
 \textbf{Instr}    & 86.06 \\
 \textbf{E}   & 87.47 \\
\textbf{Vani+Instr}   &  87.49 \\
\textbf{Vani+E}   &  88.11 \\
\textbf{Instr+E}   &  87.66 \\
\textbf{Vani+Instr+E}   &  87.92 \\
\bottomrule
\end{tabular}%
\caption{Accuracy using various types of prompts on AG's news dataset under the 16-shot setting.} \label{tab:type}
\end{table}

\begin{table*}[h]
\resizebox{\linewidth}{!}{%
\begin{tabular}{lll}
\toprule
\textbf{Dataset} & \textbf{Label Words}     & \textbf{Label Words Expansion}\\ 
\midrule
\multirow{2}{*}{\makecell[c]{SST2, IMDB\\ Yelp}} & bad  & 'bad', 'Bad', ' bad', ' Bad', 'good', 'poor', 'wrong', 'big', 'worst', ' BAD' \\
& great & 'great', 'Great', ' great', 'small', 'huge', 'little', 'good', 'perfect', 'powerful', 'big'\\ 
\midrule
\multirow{4}{*}{AG's News} & politics  & `politics', `Politics', `political', `Political', `policy', ` politics', ` Politics', `democracy', `history', `business' \\
& technology & `technology',`Technology', `tech', `software', `science', ` technology', `computer', `engineering', `technical', `device'\\ 
& business   & `business', `Business', ` business', ` Business', `commercial', `property', `biz', `office', `trade', `everything'\\ 
& sports     & `sports', `Sports', `Sport', `football', ` sports', ` Sports', `music', `gaming', `games', `military'\\
\midrule
\multirow{10}{*}{Yahoo} & soc & 'soc', 'Soc', 'social', 'community', 'offic', 'civil', ' soc', 'sn', 'mom', ' Soc'\\
& science & 'science', 'Science', 'scientific', ' science', 'technology', 'biology', 'research', 'history', 'scient', 'evidence'\\
& health &'health', 'Health', ' health', 'medical', 'healthy', ' Health', 'hospital', 'Medical', 'cancer', 'safety'\\
&education&'education', 'Education', ' education', 'learning', 'training', 'educated', 'awareness', 'student', 'college', 'school'\\
&com&'com', 'Com', ' com', 'COM', ' Com', 'org', 'comm', 'co', 'net', 'gov'\\
&sports & 'sports', 'Sports', 'Sport', 'football', ' sports', ' Sports', 'music', 'gaming', 'games', 'military'\\
&business&'business', 'Business', ' business', ' Business', 'commercial', 'property', 'biz', 'office', 'trade', 'everything'\\
&ent &'ent', 'ENT', 'ents', 'enting', 'ented', 'ant', 'ente', 'ento', 'ency', 'enta'\\
&family&'family', 'Family', ' family', 'community', ' Family', 'daughter', 'brother', 'parents', 'mom', 'small'\\
&politics & 'politics', 'Politics', 'political', 'Political', 'policy', ' politics', ' Politics', 'democracy', 'history', 'business'\\
\midrule
\multirow{14}{*}{DBPedia} & company & 'company', 'Company', ' company', 'project', 'community', 'family', 'Companies', 'city', 'business', 'office' \\
& school &  'school', 'School', ' school', 'college', ' School', 'chool', 'fashioned', 'education', 'student', 'program'\\ 
& artist &  'artist', 'Artist', ' artist', ' Artist', 'music', 'creator', 'director', 'manager', 'editor', 'student'\\ 
& ath &  'ath', 'aths', 'athe', 'ATH', 'athi', 'atha', 'athy', 'oth', 'athing', 'athan'\\ 
& politics &'politics', 'Politics', 'political', 'Political', 'policy', ' politics', ' Politics', 'democracy', 'history', 'business'\\
& trans &'trans', 'Trans', ' Trans', ' trans', 'rans', 'poly', 'simple', 'project', 'small', 'digital'\\
& building & 'building', 'build', 'builder', 'Building', 'builders', ' building', 'built', ' Building', 'making', 'fighting'\\
& river & 'river', 'River', 'rider', ' river', 'roller', 'lake', 'city', 'later', 'country', 'current'\\
& vill & 'vill', 'Vill', ' Vill', 'vell', 'ville', 'vag', 'font', 'coll', 'christ', ' vill'\\
& animal & 'animal', 'Animal', ' animal', 'species', 'human', 'monster', ' Animal', 'horse', 'baby', 'adult'\\
& plant & 'plant', ' plant', ' Plant', 'bomb', ' Plants', 'flower', 'forest', 'train', 'fruit', 'print'\\
& album & 'album', ' album', ' Album', 'music', 'movie', 'episode', 'concert', 'download', 'artist', 'film'\\
& film & 'film', 'Film', 'movie', ' film', ' Film', 'Movie', ' Films', 'music', 'video', 'picture'\\
& book & 'book', 'books', 'Book', ' book', ' Book', 'BOOK', 'Books', ' books', 'sheet', 'ebook'\\
\bottomrule
\end{tabular}
}
\caption{The results of label words expansion.}
\label{tab:expand}
\end{table*}

\begin{table*}[h]
\resizebox{\linewidth}{!}{%
\begin{tabular}{lll}
\toprule
\textbf{Dataset}        & \textbf{Template}       & \textbf{Label Words}\\ 
\midrule 
\multirow{3}{*}{SST2}          & $x$ A \texttt{[MASK]} movie. & \multirow{3}{*}{bad, great} \\
 & $x$ A \texttt{[MASK]} film. & \\
&  $x$ A \texttt{[MASK]} piece of work.\\  
\midrule
\multirow{3}{*}{Yelp}  & $x$ In summary, it was \texttt{[MASK]}. & \multirow{3}{*}{bad, great}\\
 & $x$ All in all, it was \texttt{[MASK]}. \\
 &  $x$ A \texttt{[MASK]} review.\\
\midrule
\multirow{3}{*}{IMDB}  & $x$ In summary, it was \texttt{[MASK]}. & \multirow{3}{*}{bad, great} \\
 & $x$ All in all, it was \texttt{[MASK]}. \\
&  $x$ In summary, the film was \texttt{[MASK]}.\\ 
\midrule
\multirow{3}{*}{AG's News}  & {[} Topic : \texttt{[MASK]} {]} $x_1$ $x_2$  & \multirow{3}{*}{politics, sports, business, technology} \\
 & {[} Category : \texttt{[MASK]} {]} $x_1$ $x_2$ \\
 & $x_1$ $x_2$ The topic is about \texttt{[MASK]}. \\ 
\midrule
\multirow{3}{*}{Yahoo}  & {[} Topic : \texttt{[MASK]} {]} $x_1$ $x_2$  & \multirow{3}{*}{\makecell[l]{society, science, health, education,\\ computers, sports, business,\\ entertainment, family, politics} } \\
& A \texttt{[MASK]} question : $x_1$ $x_2$ \\
& $x_1$ $x_2$ The topic is about \texttt{[MASK]}.\\
\midrule
\multirow{3}{*}{DBPedia}  & {[} Topic : \texttt{[MASK]} {]} $x_1$ $x_2$  & \multirow{3}{*}{\makecell[l]{company, school, artist, athlete, politics,\\ transportation, building, river, village,\\ animal, plant, album, film, book}} \\
& $x_1$ $x_2$ $x_1$ is a \texttt{[MASK]}. \\
&  $x_1$ $x_2$ In this sentence, $x_1$ is a \texttt{[MASK]}. \\ 
\midrule
RTE & $x_1$? \texttt{[MASK]}, $x_2$              & No, Yes      \\  
SNLI &    $x_1$. \texttt{[MASK]}, $x_2$  & No, Maybe, Yes       \\ 
MNLI-m/mm &  $x_1$! \texttt{[MASK]}, $x_2$  &  No, Maybe, Yes  \\
\bottomrule
\end{tabular}
}
\caption{The templates and label words used in our experiments. Different datasets use different forms of input, such as sentence$x$ and sentence pairs $x_1$ $x_2$.}
\label{tab:template}
\end{table*}

\begin{table*} \setlength{\tabcolsep}{8pt}
\centering
\begin{tabular}{lcccccc}
\toprule
\textbf{Template Id}&\textbf{Template}\\
\midrule
1& {[} Topic : \texttt{[MASK]} {]} $x_1$ $x_2$  \\
2& {[} Category : \texttt{[MASK]} {]} $x_1$ $x_2$\\
3& $x_1$ $x_2$ The topic is about \texttt{[MASK]}.\\
4& \texttt{[MASK]} Alert Blog Dialogue Diary Accountability $x_1$ $x_2$\\
5& A \texttt{[MASK]} news : $x_1$ $x_2$\\
\bottomrule
\end{tabular}
\caption{Five templates used in our experiments. The fourth prompt comes from \citet{DBLP:conf/emnlp/DengWHWGSSXH22}.}  \label{tab:effects_of_prompt}
\end{table*}

\begin{table*} \setlength{\tabcolsep}{8pt}
\centering
\begin{tabular}{lcccccc}
\toprule
\textbf{Template Type}&\textbf{Template}\\
\midrule
Vani& {[} Topic : \texttt{[MASK]} {]} $x_1$ $x_2$\\
\midrule
\multirow{2}{*}{Instr} & \multirow{2}{*}{\makecell[l]{Classify the news articles into the categories of Politics, Sports, Business, and Technology.\\ $x_1$ $x_2$ This topic is about \texttt{[MASK]}.}} \\
\\
\midrule
\multirow{3}{*}{E} & \multirow{3}{*}{\makecell[l]{Article: Wall St. Bears Claw Back Into the Black.
(Reuters), Reuters - Short-sellers,\\ Wall Street's dwindling of ultra-cynics, are seeing green again. Answer: Business.\\ Article: $x_1$ $x_2$ Answer: \texttt{[MASK]}.}}\\
\\
\\
\bottomrule
\end{tabular}
\caption{Three different types of templates (i.e., vanilla prompt, instruction prompt, and example prompt) used in our experiments.}  \label{tab:temp_ag}
\end{table*}

\section{Effectiveness on Various Types of Prompts}
In this section, we further show the effectiveness on various types of prompts, including instruction and example.
Specifically, we use the vanilla prompt 1 in Figure \ref{fig:tempalte} (denoted as Vani), instruction (denoted as Instr), and example (denoted as E, we use the first training data from the AG‘s news dataset to construct example prompt) on the AG's news dataset under the 16-shot setting.
Three prompts as shown in Table \ref{tab:temp_ag}.

Firstly, our method usually achieves performance improvement compared to using a single prompt in Table~\ref{tab:type}.
Specifically, using two prompts exceeds using either prompt individually.
For example, Vani+Instr exceeds Vani and Instr, respectively.
It is worth noting that Vani+E exceeds all results in Figure \ref{fig:tempalte}.
This shows that using two different types of prompts can sometimes yield better results.
Secondly, the performance fluctuation of using two prompts is smaller than that of a single prompt. Thirdly, Vani+Instr+E does not exceed Vani+Instr due to the poor performance of Instr.
This is because RoBERTa-large has not been fine-tuned by instruction-tuning.

\renewcommand{\algorithmicensure}{\textbf{Output:}}
\renewcommand{\algorithmicrequire}{\textbf{Input:}}
\begin{algorithm}[b]
\caption{The training flow of our method} \label{alg:train}
\label{Algorithm}
    \begin{algorithmic}[1]
     \Require Training set and PLM $\mathcal{M}$.
    \Ensure The parameters of linear layer $\mathbf{W}$ and prototypes $\mathbf{R}$
    \State{\textcolor{blue}{$\rhd$ Extract features and initialize class prototypes:}}
    \State {Randomly initialize class prototypes $\bm{R}$}
    \State{Obtain a feature set $\bm{H}$ via PLM $\mathcal{M}$}
    \State{\textcolor{blue}{$\rhd$ Training:}}
    \For {epoch $=1, 2, \dots, T$}
        \State{Sample a data $x_i$ and use linear layer to get the text representations $\bm{V}_i$}
        \State{\textcolor{blue}{$\rhd$ Computing OT scores via Sinkhorn's algorithm:}}
        \For {$k=1,2,\dots,N$}
        \State{Calculate the cost matrix $\bm{C}_k=\bm{1}-\bm{V}\bm{R}_k^\top$ of each class}
        \For{$t_{in}=1,2,\dots,T_{in}$}
        \State{$Z = \exp(-\bm{C}/\lambda)$}
        \State{$\bm{u}^{(t_{in})} =\bm{u}/(Z\bm{v}^{(t_{in}-1)}) $}
        \State{$\bm{v}^{(t_{in})} =\bm{v}/(Z^\top\bm{u}^{(t_{in})}) $}
        \State{$\Delta_{v} =\sum |\bm{v}^{(t_{in})}- \bm{v}^{(t_{in}-1)}|/N $}
        \If{$\Delta_{v}< \delta$}
        \State{break}
        \EndIf
        \EndFor
        \State{Obtain optimal transport plan  as $\bm{T}^*_k= \text{diag}(\bm{u}^{(t)}) \exp(-\bm{C}_k/\lambda)\text{diag}(\bm{v}^{(t)})$}
        \State{Calculate the OT scores by Eq. (2)}
        \EndFor
        \State{Update the parameters with cross-entropy loss by Eq. (3)}
    \EndFor
    \State \Return The well-trained linear layer $\mathbf{W}$ and prototypes $\mathbf{R}$
    \end{algorithmic}
\end{algorithm}

\end{document}